%% file: main.tex
\documentclass[runningheads]{llncs}

\usepackage{eccv}

\usepackage{eccvabbrv}

\usepackage{graphicx}
\usepackage{booktabs}

\usepackage[accsupp]{axessibility}  % Improves PDF readability for those with disabilities.
\usepackage{bigstrut,multirow,rotating}
\usepackage{array}
\newcolumntype{C}[1]{>{\centering\arraybackslash}p{#1}}
\usepackage{colortbl}
\usepackage{booktabs} % for better-looking tables
\usepackage{wrapfig}
\usepackage{makecell}

\usepackage{xcolor}
\newcommand{\jxnote}[1]{\textcolor{black}{#1}}
\newcommand{\crnote}[1]{\textcolor{black}{#1}}

\usepackage[linesnumbered,ruled,vlined]{algorithm2e}

\usepackage{hyperref}
\usepackage{authblk}

\usepackage{orcidlink}
\newcommand{\equalcontrib}{\textsuperscript{*}}
\newcommand{\correspondingauthor}{\textsuperscript{\textdagger}}

\begin{document}

\title{Iterative Ensemble Training with Anti-Gradient Control for Mitigating Memorization in Diffusion Models } 

\titlerunning{IET-AGC: Mitigating Memorization in Diffusion Models}

\author{
Xiao Liu\inst{1}\equalcontrib\orcidlink{0009-0004-3237-6712}\and
Xiaoliu Guan\inst{1}\equalcontrib\orcidlink{0009-0000-3678-4255} \and
Yu Wu\inst{1}\correspondingauthor\orcidlink{0000-0002-1680-8253} \and
Jiaxu Miao\inst{2}\correspondingauthor\orcidlink{0000-0002-4238-8475}
}

\authorrunning{X.~Liu et al.}


\institute{School of Computer Science, Wuhan University \email{\{xiaoliu,liuxiaoguan,wuyucs\}@whu.edu.cn} 
\\
\and
School of Cyber Science and Technology, Sun Yat-sen University \email{miaojx@mail.sysu.edu.cn}}

\maketitle
% 添加脚注

\begin{abstract}
  % The abstract should summarize the contents of the paper. 
  % LNCS guidelines indicate it should be at least 70 and at most 150 words.
  % Please include keywords as in the example below. 
  % This is required for papers in LNCS proceedings.

 % diffusion models, known for their ability to generate novel and high-quality samples, have recently raised concerns due to their data memorization behavior, which poses privacy risks. In this study, we propose a novel approach Iterative Ensemble Training with Anti-Gradient Control (IET-AGC) to address this issue by introducing two modules aimed at reducing the memory capacity of diffusion models. The first module focuses on iterative interactive training of submodels, while the second module employs a gradient control technique to prevent overfitting and reduce memorization. We conduct extensive experiments and analysis on three datasets to evaluate the effectiveness of our method. Our results show that our approach successfully reduces memory capacity while even improving model performance slightly, highlighting its potential for enhancing privacy in diffusion models.

\jxnote{Diffusion models, known for their tremendous ability to generate novel and high-quality samples, have recently raised concerns due to their data memorization behavior, which poses privacy risks. Recent approaches for memory mitigation either only focused on the text modality problem in cross-modal generation tasks or utilized data augmentation strategies. In this paper, we propose a novel training framework for diffusion models from the perspective of visual modality, which is more generic and fundamental for mitigating memorization. To facilitate ``forgetting'' of stored information in diffusion model parameters, we propose an iterative ensemble training strategy by splitting the data into multiple shards for training multiple models and intermittently aggregating these model parameters. Moreover,  practical analysis of losses illustrates that the training loss for easily memorable images tends to be obviously lower. Thus, we propose an anti-gradient control method to exclude the sample with a lower loss value from the current mini-batch to avoid memorizing. Extensive experiments and analysis on \crnote{four} datasets are conducted to illustrate the effectiveness of our method, and results show that our method successfully reduces memory capacity while even improving the performance slightly. Moreover, to save the computing cost, we successfully apply our method to fine-tune the well-trained diffusion models by limited epochs, demonstrating the applicability of our method. Code is available in \url{https://github.com/liuxiao-guan/IET_AGC}.}

\end{abstract}
\renewcommand{\thefootnote}{\fnsymbol{footnote}}
\footnotetext[1]{Equal contribution.}
\renewcommand{\thefootnote}{\textdagger}
\footnotetext[2]{Corresponding author.}
\section{Introduction}
\label{sec:intro}
\input{introduction.tex}

\section{Related Work}

\subsubsection{Memorization in Generative Models.}
Several studies have examined the memorization capabilities of the generative model. 
Generative Adversarial Networks (GANs)~\cite{goodfellow2020generative} have been at the forefront of this research area. 
% The interplay between memorization and GANs has been a focal point of several studies. 
As Webster \etal~\cite{webster2021person} demonstrated when applied to face datasets, GANs can occasionally replicate.
% Feng \etal~\cite{feng2021gans}investigated conditions causing GANs to replicate training data, found pixel-level copies, and noted an inverse relationship with dataset complexity and size.
Prior study~\cite{carlini2021extracting} explored an adversarial attack on language models like GPT-2~\cite{radford2019language}, where individual training examples can be recovered, including personally identifiable information and unique text sequences.

Recent studies have shifted their attention toward diffusion models. 
Somepalli \etal~\cite{somepalli2023diffusion} found that diffusion models accurately recall and replicate training images, especially noted with models like the Stable Diffusion model~\cite{rombach2022high}.
% Somepalli \etal~\cite{somepalli2023diffusion} discovered that diffusion models could recollect and replicate their training images with incredible accuracy. 
% It also pinpointed instances where diffusion models, such as the Stable diffusion model~\cite{rombach2022high}, clearly reproduce from their training data.
Building upon this discovery, Carlini \etal~\cite{carlini2023extracting} developed a tailored black-box attack for diffusion models. They generated images and implemented a membership inference attack to assess density.
Webster \etal~\cite{webster2023reproducible} demonstrated a more efficient extraction attack with fewer network evaluations, identified "template verbatims," and discussed its persistence in newer systems. 
Recent research has shifted towards exploring the theoretical aspects of memory in diffusion models.
% ~\cite{yoon2023diffusion,kadkhodaie2023generalization} investigated the interrelationship between the phenomena of memory and generalization. 
~\cite{yoon2023diffusion} discovered that generalization and memorization are mutually exclusive occurrences and further demonstrated that the dichotomy between memorization and generalization can be apparent at the class level.
Gu \etal~\cite{gu2023memorization} extensively studied how factors like data dimension, model size, time embedding, and class conditions affect the memory capacity of the diffusion model.

\subsubsection{Memorization Mitigation.} 
% Out of consideration for privacy and copyright protection, some work has proposed solutions to the memorization issues of diffusion models. 
The mitigation measures have primarily been concerned with filtering inputs and deduplication. 
For example, Stable Diffusion employed well-trained detectors to identify unsuitable generated content. 
However, these temporary solutions can be easily bypassed~\cite{wen2024hard,rando2022red} and do not effectively prevent or lessen copying behavior on a broad scale. 
% ~\cite{kumari2023ablating,gandikota2023erasing,huang2023receler,schramowski2023safe} deleted entire concepts such as copyrighted concepts, licensed images, and personal photos from the model.
Kumari \etal~\cite{kumari2023ablating} designed an algorithm to align the image distribution with a specific style, instance, or text prompt they aim to remove, to the distribution related to a core concept. 
This stopped the model from producing target concepts based on its text condition.
However, these approaches are inefficient because they necessitate a list of all concepts to be erased, and have not addressed the key issue of how to reduce the memory capacity of the model.
~\cite{dockhorn2022differentially,ghalebikesabi2023differentially} explored the use of differential privacy (DP)~\cite{dwork2006differential} to train diffusion models or fine-tune ImageNet pre-trained models. However, their focus was on ensuring the privacy of the training of diffusion models, not on the privacy of the images generated by the diffusion models. 
Daras \etal~\cite{daras2024ambient} introduced a technique for training diffusion models utilizing tainted data. By incorporating additional corruption before applying noise, their methodology prevents the model from overfitting to the training data. But their training requires a considerable amount of time. 
~\cite{somepalli2024understanding,wen2023detecting} also suggested a series of recommendations to mitigate copying such as randomly replacing the caption of an image with a random sequence of words, but most of which are limited to text-to-image models. Our work focuses on the nature of memorization in diffusion models, especially for unconditional ones. 

\section{Method}
In this section, we present our methodology aimed at mitigating the memorization in diffusion models, without sacrificing excessive image quality.

\begin{figure}[tb]
  \centering
  \setlength{\abovecaptionskip}{7pt} % 设置标题上方的间距为 -5pt
  \setlength{\belowcaptionskip}{-14.5pt} % 设置标题下方的间距为 -5pt
  \includegraphics[width=1.0\linewidth]{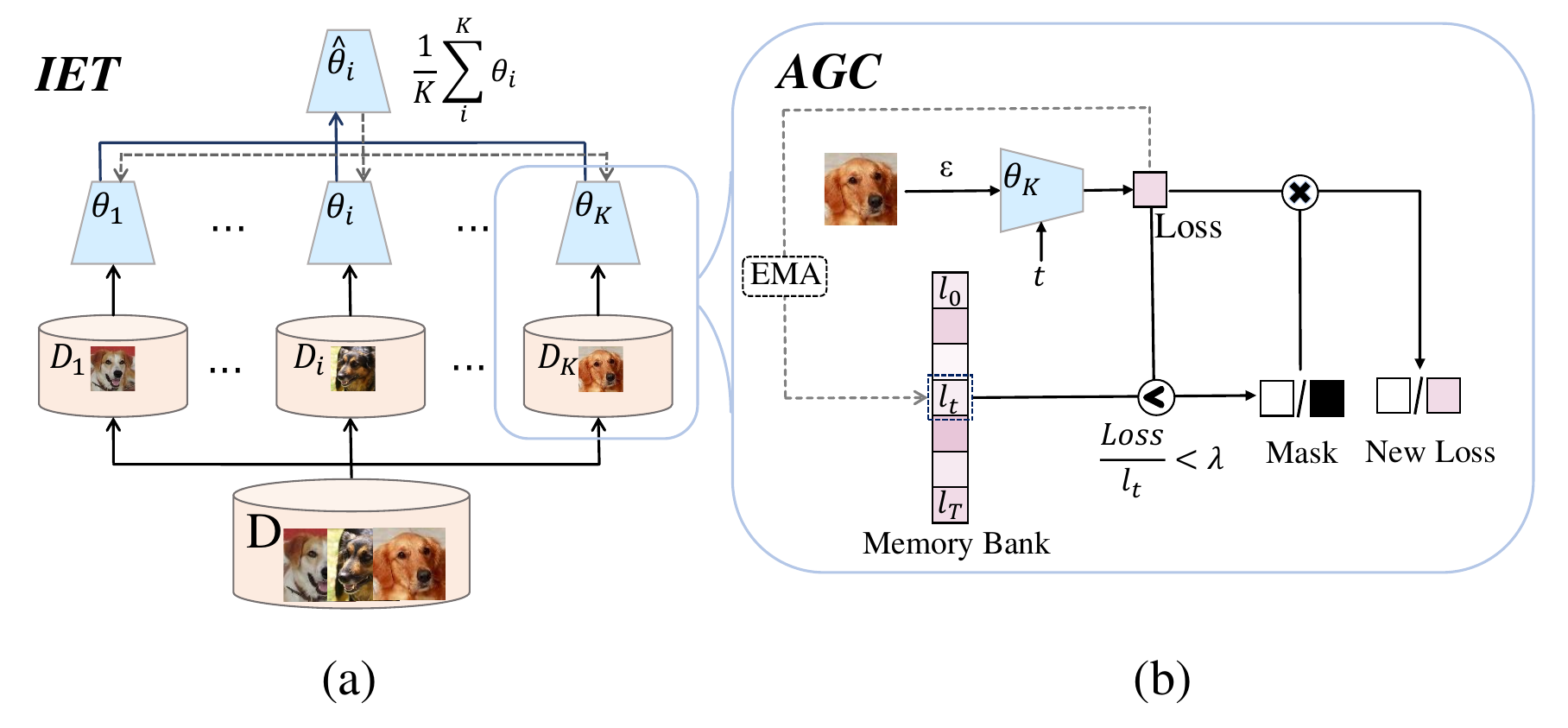}
  \caption{Overview of our IET-AGC method. \textbf{(a)} Iterative Ensemble Training (IET): we divide the dataset $D$ into $K$ different data shards. Each shard $D_i$ trains a separate diffusion model $\theta_i$. After a period of training, these models are merged by averaging and repeating this training strategy.
  \textbf{(b)} Anti-Gradient Control (AGC): during the training process, we dynamically update and maintain a memory bank of losses at each time step.
  For loss values smaller than $\lambda$ times the corresponding memory bank, we exclude these losses to prevent the model from memorizing such images.}
  \label{fig:method}
\end{figure}

\subsection{Iterative Ensemble Training}
\jxnote{Training data are stored in parameters of diffusion models due to over-optimization, and the model ensemble strategy aggregates parameters to re-organize model knowledge. Thus,}  to mitigate the memorization, we propose a method that trains multiple diffusion models on different data shards of a dataset, merges them after a certain period, and then repeats these two stages iteratively.

\subsubsection{Training on Different Data Shards.}Unlike the training methods of previous diffusion models, which train a single model on the entire dataset once. In this paper, we divide a dataset into an equal number of shards and then train the corresponding diffusion models on each separate part. 
 If the dataset contains class information, we divide it equally along the class dimension, so that each data shard contains all classes, with the same number of instances for each class. 

Specifically, suppose the dataset $D$ contains $C$ classes, with each class having $N_c$ samples, and the total number of samples is $N=\sum_{c=1}^{C}N_c$. We divide the dataset into $K$ equal parts, with each part containing $\frac{N}{K}$ samples. So the $i$th part of the dataset $D_i$ is represented as
\begin{equation}
 D_i = \bigcup_{c=1}^{C} \{ (x_{c, i, k}, y_{c, i, k}) \mid k=1, 2, \ldots, \frac{N_c}{K} \},
 \label{eq:divided_datasets}
\end{equation}
where $(x,y)$ represents the sample and its corresponding label. \crnote{Notably, $C=1$ indicates that the dataset does not contain class information.} Then, each part $i$ trains a separate diffusion model $\theta_{i}$ scaled by the learning rate $\eta$ on its own dataset,
\begin{equation}
\theta_{i}\leftarrow \theta_{i} - \eta \nabla \mathcal{L}(\theta_{i}).
  \label{eq:local_update}
\end{equation}

\subsubsection{Merging the Multiple Diffusion Models.} After training for some time, each part obtains a different diffusion model. We simply average the weights of all models $\theta_{i}$ to obtain a global model $\widehat{\theta}$ as
\begin{equation}
\widehat{\theta}\leftarrow \frac{1}{K}\sum_{i=1}^{K}\theta_{i}.
  \label{eq:average}
\end{equation}

Then, we repeat the two stages of training on separate shards of the data and merging models, where we use the obtained global model as the initial model for the first stage.
\crnote{As each shard contains only $\frac{1}{K}$ of the total data, the training time for each model is proportionally reduced,
maintaining the overall computational cost nearly constant compared to training a single model on the entire dataset. 
The only additional computational cost comes from the periodic merging of models, which is minimal.
}
% \crnote{Although IET is inspired by federated learning,  the motivation for splitting the data and aggregating the models is to reduce the Diffusion Models' memorization of the training data. 
% In other words,
% the problem our method aims to address is entirely different from federated learning. Federated learning aims to prevent local data leakage across clients \textbf{during distributed training}. While our method aims to prevent pretrained diffusion models from generating the training data \textbf{during inference}, i.e., the memorization problem of diffusion models.}

\subsection{Loss Analysis}
To further reduce memorization of training data, we delve deeper into the causes of memorization phenomena, specifically analyzing it through the lens of the loss.
We begin by establishing the fundamental notation linked with diffusion models.
Diffusion models \cite{ho2020denoising} originate from the non-equilibrium statistical physics \cite{sohl2015deep}.
They are essentially straightforward: they operate as image denoisers.
During the training process, when given a clean image $x$, time-step $t$ is sampled from the interval [$0$, $T$], along with a Gaussian noise vector $\epsilon \sim \mathit{N} (0, I)$,
resulting in a noised image $x_t$:
\begin{equation}
  x_t = \sqrt{\alpha_t}x + \sqrt{1-\alpha_t} \epsilon, 
  \label{eq:noised_data}
\end{equation}
where the scheduled variance $\alpha_t$ varies between $0$ and $1$, with $\alpha_0 = 1$ and $\alpha_T = 0$. 
The diffusion model then removes the noise to reconstruct the original image $x$ by predicting the noise introduced, achieved through stochastic minimization of the objective function
$\frac{1}{N} \sum_{i} \mathbb{E}_{t,\epsilon} \mathcal{L} (x_i, t, \epsilon; \theta)$, where
\begin{equation}
  \mathcal{L} (x_i, t, \epsilon; \theta) = \| \epsilon - \epsilon_\theta(\sqrt{\alpha_t}x_i + \sqrt{1-\alpha_t}\epsilon, t) \|_2^2.
  \label{eq:loss}
\end{equation}

\begin{figure}[tb]
  \centering
  \setlength{\abovecaptionskip}{0pt} % 设置标题上方的间距为 -5pt
  \setlength{\belowcaptionskip}{-13pt} % 设置标题下方的间距为 -5pt
  \includegraphics[height=6.5cm]{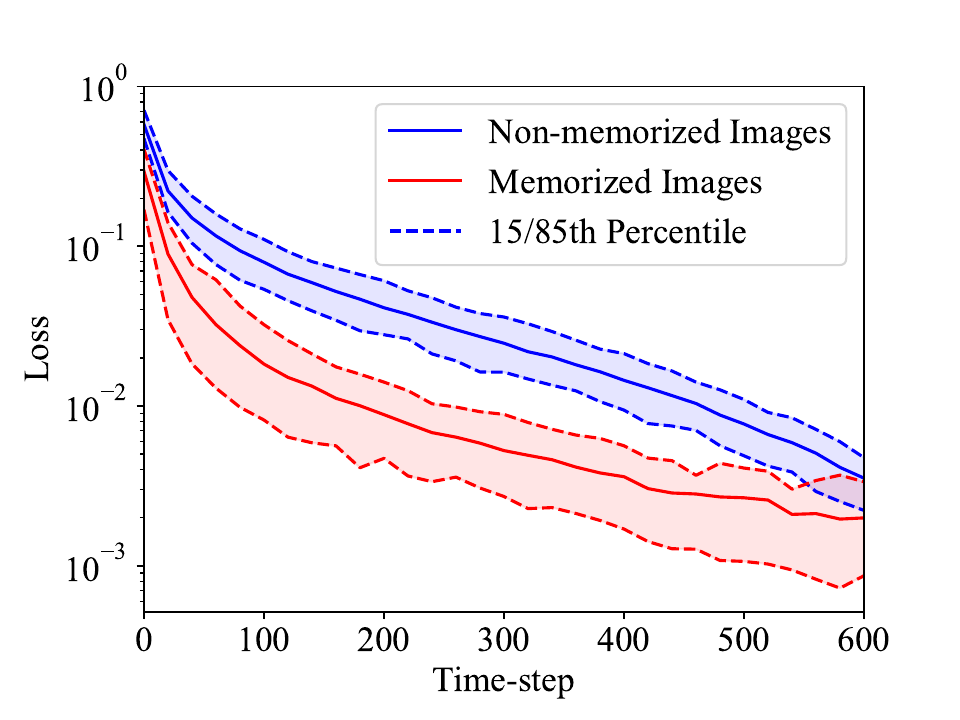}
  \caption{
  Comparison of the losses between memorized and non-memorized images. The solid line represents the averaged losses of memorized and non-memorized images, while the two dashed lines represent the losses of the 15th and 85th percentile data respectively.
  }
  \label{fig:LossAnalysis}
\end{figure}

To analyze the correlation between losses and image memorization, we identify images memorized on CIFAR-10 and calculate their loss functions at each time step. 
Similarly, we sample 256 non-memorized images from the remaining training data and compute their losses at each time step.
%Then we randomly sample 256 images from the training set to compute their loss functions in the same way, which we consider as non-memorized images.
\cref{fig:LossAnalysis} shows the comparisons of the losses when the time step is in the interval [0, 600] $(T=1000)$. 
Memorized images exhibit significantly smaller loss values during this period, indicating that the model tends to reconstruct noise into such images.

\subsection{Anti-Gradient Control}
In this subsection, we elaborate on how to utilize the aforementioned loss analysis to devise a training strategy aimed at alleviating the occurrence of memorization.

\subsubsection{Memory Bank.}
In order to identify images with exceptionally low loss values that are prone to memorization during training, we need to maintain the average losses for each time step. 
However, computing the average loss at each time step entails substantial computational expenses, as it necessitates evaluating the losses for all images using the model at each time step. Thus, we propose a memory bank to store and update losses during mini-batch training without increasing the time cost. However, 
the losses generally decrease with the training step growing. Therefore, when calculating the average loss in the memory bank, instead of directly averaging all losses at this time step, the aggregating weights should be higher for losses closer to the current update.
% As the diffusion model trains, the loss generally decreases. 

% To estimate the loss of the model at each time step without increasing the time cost, we introduce the concept of the memory bank.
% As the diffusion model trains, the loss generally decreases. 
% Therefore, we aim for the loss value closer to the current update to carry greater weight in the calculation of the average. 
Specifically, we initialize an array of length $T$ with zeros, termed the memory bank. 
The initialization values of the memory bank seem insignificant.
After calculating the loss for a mini-batch, we update the memory bank using the Exponential Moving Average (EMA) method based on the loss and the sampled time step, thereby better reflecting the current state of the model:
\begin{equation}
  l_{t} \leftarrow \gamma \cdot l_{t} + (1 - \gamma) \cdot \mathcal{L} (x, t, \epsilon; \theta),
  \label{eq:ema}
\end{equation}
where $\gamma$ represents the smoothing factor, and $l_{t}$ represents the averaged loss in the memory bank at time step $t$. 

\subsubsection{Anti-Gradient Control (AGC).}
In previous observations, if the model exhibits memorization of a certain sample, the loss value of the model on that sample tends to be abnormally small.
Thus, we use the ratio of the training loss of a certain sample to the mean loss in the memory bank at the time 
 step $t$ as a measure to detect memorization:
\begin{equation}
  r = \frac{\mathcal{L} (x, t, \epsilon; \theta)}{l_t}.
  \label{eq:ratio}
\end{equation}
A smaller ratio of $r$ indicates a higher likelihood of the image being memorized. 
To facilitate this determination, we can establish a configurable threshold denoted as $\lambda$. 
If the loss ratio $r$ falls below this threshold $\lambda$, the image is classified as memorized.
At this point, we generate a mask that sets the loss value corresponding to this image to zero, \ie skipping this image in the mini-batch, as shown in the following function,
\begin{equation}
\mathcal{L}(x, t, \epsilon; \theta) = 
\begin{cases} 
0 & \text{if } r < \lambda \\
\mathcal{L}(x, t, \epsilon; \theta) & \text{otherwise}.
\end{cases}
\label{eq:update_loss}
\end{equation}
Since the model has encountered the sample during training, excluding it is unlikely to have a significant impact on the model's performance.

\section{Experiments}
% \label{sec:blind}
% We now present the empirical results of our method. 
% \crnote{
% In \cref{sec:datasets}, we train diffusion models from scratch to demonstrate the effectiveness of our approach in reducing model memorization.
% }
% \crnote{
% In \cref{sec:finetune}, we fine-tune pre-trained visual-only diffusion model DDPM~\cite{ho2020denoising} and text-conditioned diffusion model Stable Diffusion ~\cite{rombach2022high}, demonstrating our approach's ability to make models forget memorized images without compromising the quality of generated images.
% }
% % In \cref{sec:datasets}, we introduce the datasets we conduct experiments on.
% % In \cref{sec:setup}, we present the implementation details including our training and memorized image-extracting details.  
% % In \cref{sec:result}, we show the results on datasets CIFAR-10~\cite{krizhevsky2009learning}, CIFAR-100~\cite{krizhevsky2009learning} and AFHQ-DOG~\cite{choi2020stargan}. 
% In \cref{sec:analysis}, we analyze skipping.
% In \cref{sec:ablation}, we conduct ablation studies to verify the necessity of each component of our method. 

% \subsection{Training Diffusion Models from Scratch}

\subsection{Experimental Setup}
\label{sec:datasets}
\crnote{
\textbf{Datasets.}
We evaluate our method on CIFAR-10~\cite{krizhevsky2009learning}, CIFAR-100~\cite{krizhevsky2009learning}, AFHQ-DOG~\cite{choi2020stargan} for unconditional generation, and LAION-10k~\cite{somepalli2024understanding} for text-conditioned generation. 
CIFAR-10 and CIFAR-100 consist of 50,000 32x32 color images, divided into 10 and 100 classes, respectively. 
AFHQ-DOG is a subset of the AFHQ dataset with approximately 5,000 512x512 dog images, resized to 64x64 for our experiments.
LAION-10k is a subset of LAION~\cite{schuhmann2021laion}, comprising 10,000 image-text pairs with each image having a resolution of 256x256 pixels.
}
% We evaluate the effectiveness of our method on CIFAR-10~\cite{krizhevsky2009learning}, CIFAR-100~\cite{krizhevsky2009learning}, and AFHQ-DOG~\cite{choi2020stargan}. CIFAR-10 is a dataset that consists of 50,000 32x32 color images, which are split into 10 different classes, with each class containing 5,000 images for training. Similarly, CIFAR-100 is a more extensive dataset that includes 50,000 32x32 color images. These images are split into 100 classes, with each class containing 500 training images. For AFHQ-DOG, it is a subset of the AFHQ dataset which is a high-quality dataset of animal faces. It comprises approximately 5,000 images, each with a resolution of 512x512 pixels. In our experiment, we resize these images to a resolution of 64x64 pixels.

% \subsection{Experimental Setup}
\label{sec:setup}
% \subsubsection{Implementation Details.}
% \subsection{Implementation Details}
% Our implementation details contain two parts: training and memorized image extracting attack. 

\textbf{Implementation Details of Training.} 
We conduct experiments on training unconditional diffusion models from scratch using the CIFAR-10, CIFAR-100, and AFHQ-DOG datasets. 
The IET framework divides CIFAR datasets into 10 shards and AFHQ-DOG into 5 shards.
Threshold $\lambda$ is set to 0.5 for CIFAR datasets and 0.714 for AFHQ-DOG. 
% For CIFAR-10 and CIFAR-100, we set the batch size to 128 and trained for 400k and 580k iterations, respectively. For AFHQ-DOG, the batch size was 60, with 180k iterations.
Additionally, we perform experiments on fine-tuning pre-trained diffusion models using the CIFAR-10 dataset, maintaining the same hyperparameters as those used for training from scratch. 
% We find that fine-tuning for just two epochs was sufficient to mitigate the model's memory effect. 
% For text-conditioned diffusion models, we follow the setup of A, setting the batch size to 16 and training for 100k iterations.
\crnote{
To demonstrate the effectiveness of our method in text-conditioned diffusion models, we fine-tune Stable Diffusion on LAION-10k.
The IET framework divides the LAION-10k dataset into 8 shards, with the threshold $\lambda$ set to 0.8.}
The smoothing factor $\gamma$ is 0.8 for all datasets.
Further details are in the supplementary material.

\textbf{Extracting Memorized Image.} 
% We adopt the detection rule introduced in Carlini \etal~\cite{carlini2023extracting} which considered $\bar{x}$  as memorized if the $\ell_{2}$ distance to its nearest neighbor in the training set is much lower compared to the $n$ closest neighbors. We slightly modify our attack to use the distance:
We adopt Carlini's detection rule~\cite{carlini2023extracting} for unconditional generation, considering 
$\bar{x}$ as memorized if the $\ell_{2}$ distance to its nearest neighbor is significantly lower compared to the $n$ closest neighbors. We modify this rule to:
\begin{equation}
\ell_{2}(\bar{x},x;\mathbb{S}^n_{\bar{x}}) = \frac{\ell_{2}(\bar{x},x)}{\mathbb{E}_{y\in{\mathbb{S}^n_{\bar{x}}}}[\ell_{2}(\bar{x},y)]},
  \label{eq:attack}
\end{equation}
% where we set n as 50, and $\mathbb{S}^n_{\bar{x}}$ consists of  $n$ elements from the training dataset that is closest to the example $\bar{x}$. We can turn this into a binary classifier by threshing its values with $\delta_{V}$ as follows
where $n = 50$ in our experiment. A binary classifier is defined as:
\begin{equation}
    IsMemo(\bar{x},x;\mathbb{S}^n_{\bar{x}},\delta_{V}) = \mathrm{ \boldmath{1}}_{\ell_{2}(\bar{x},x;\mathbb{S}^n_{\bar{x}})\leq\delta_{V}}.
  \label{eq:attack_threshold}
\end{equation}
The more images below $\delta_{V}$, the stronger the model's memorization.
We generate 65,536 images per model, calculate their $\ell_{2}$ distances, and count images below thresholds $\delta_{V}$ of 0.4, 0.5, and 0.6 to quantitatively evaluate the model's memorization, denoted as MQ$_{0.4}$, MQ$_{0.5}$ and MQ$_{0.6}$.

We adopt Somepalli's detection rule~\cite{somepalli2024understanding} for text-conditioned generation, quantify memorization using a similarity score derived from the dot product of SSCD features ~\cite{pizzi2022self} of $\bar{x}$ and the nearest neighbor $n_0$:
\begin{equation}
    \zeta = E(\bar{x})^T \cdot E(n_0)
\end{equation}
where $E(\cdot)$ is SSCD~\cite{pizzi2022self}. The dataset similarity score is then defined as the 95th percentile of this distribution.
% To be specified, for each model, we generate 65,536 images, calculate their $\ell_{2}$ distance to each image in the training set and then count the number of images below the threshold, which we refer to as memorized quantity, abbreviated as \textbf{MQ}.
% In the experiments, we choose $\delta_{V}$ as 0.4, 0.5, and 0.6 to serve as loose or strict standards for image similarity. 
% The corresponding metrics are denoted as MQ$_{0.4}$, MQ$_{0.5}$ and MQ$_{0.6}$ respectively.
% This allows us to verify the stability of our method. 

\subsection{Experimental Results}
\label{sec:result}
\textbf{Training from Scratch.}
The experimental results of our method and four competitive methods are shown in Tab.~\ref{tab:attack_result}.
"Default" denotes the conventional training approach of DDPM~\cite{ho2020denoising}. 
``DP-SGD'' denotes the method of Differentially Private Stochastic Gradient Descent~\cite{abadi2016deep}, which involves clipping and adding noise to the model's gradients to protect privacy, albeit at the cost of some image quality.
Carlina \etal~\cite{carlini2023extracting} found that utilizing DP-SGD could lead to consistent model divergence. 
To make DP training more stable, we modify the amplitude of the added noise to be the product of the gradient norm and the noise multiplier:
\begin{equation}
    \sigma = ||\nabla \mathcal{L}|| \times \tau,
  \label{eq:dp}
\end{equation}
where $\tau$ represents the noise multiplier. 
% In this way, the noise size dynamically adjusts with the gradient size, making the training process more robust.
``Adding noise'' denotes a method of directly adding Gaussian noise to the images during training, with a mean of 0 and a variance of 0.1.
``Ambient Diffusion''~\cite{daras2024ambient} protects privacy by training generative models on highly corrupted samples, preventing the model from observing clean training data.

% We further highlight the effectiveness of our method by contrasting it with DP-SGD (Differentially Private Stochastic Gradient Descent)~\cite{abadi2016deep}, a technique that involves the clipping and noising of the model's gradients to safeguard privacy by preventing substantial data leakage about the inclusion of any individual image in the dataset, albeit at the cost of some image quality.
% In a study by Carlina \etal~\cite{carlini2023extracting}, they found that utilizing DP-SGD could lead to consistent model divergence. 
% To make DP training more stable, we modify the amplitude of the added noise to be the product of the gradient norm and the noise multiplier.
% In other words, for each parameter update, the standard deviation of the random noise 
% we add to the gradient is calculated as: 
% \begin{equation}
%     \sigma = ||\nabla \mathcal{L}|| \times \tau
%   \label{eq:dp}
% \end{equation}
% where $\tau$ represents the noise multiplier. In this way, the size of the noise will dynamically adjust with the size of the gradient, making the training process more robust.

% We also propose a method of directly adding noise to the data.
% Specifically, we add Gaussian noise to the images during training, with a mean of 0 and a variance of 0.1.
% Results are reported in ~\cref{tab:attack_result}. 

Results in Tab.~\ref{tab:attack_result} show that adding noise to the training images or gradients reduces the quality of the generated images. 
However, it still does not resolve the issue of training image memorization.
Despite Ambient Diffusion also reducing memorization, it leads to a significant increase in FID (from 8.81 to 11.7), indicating a notable degradation of image quality.
Compared with the default training approach, our method maintains or even slightly improves the generative quality by reducing the FID score.
At the same time, our method significantly reduces the diffusion model's memorization of the training data. As shown in Tab.~\ref{tab:attack_result}, for the MQ$_{0.4}$ score, the number of memorized images reduced by 87.3\%, 66.4\%, and 85.3\% compared with the default training on CIFAR-10, CIFAR-100 and AFHQ-DOG, respectively, illustrating the effectiveness of our method.

% Adding noise directly to the data did not reduce the memorization level of these images by the model.
% Although this significantly degrades the quality of the generated images, the model can still remember these images. Similarly, adding noise to the gradient does not reduce the model's memory capacity for these images. However, on all three datasets, our method successfully slightly decreases the FID while also reducing the number of images remembered by the model. This result validates the effectiveness of IET-AGC.

\begin{figure}[tb]
  \setlength{\abovecaptionskip}{22pt} % 设置标题上方的间距为 -5pt
  \setlength{\belowcaptionskip}{-10pt} % 设置标题下方的间距为 -5pt
  \centering
  \begin{subfigure}{0.4\linewidth}
    \includegraphics[width=1.0\linewidth]{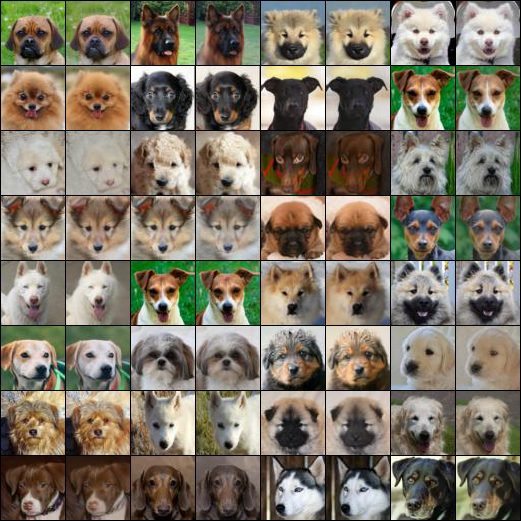}
    \caption{Default Training}
    \label{fig:most}
  \end{subfigure}
  % \hspace{0.1\linewidth}
  \begin{subfigure}{0.4\linewidth}
    \includegraphics[width=1.0\linewidth]{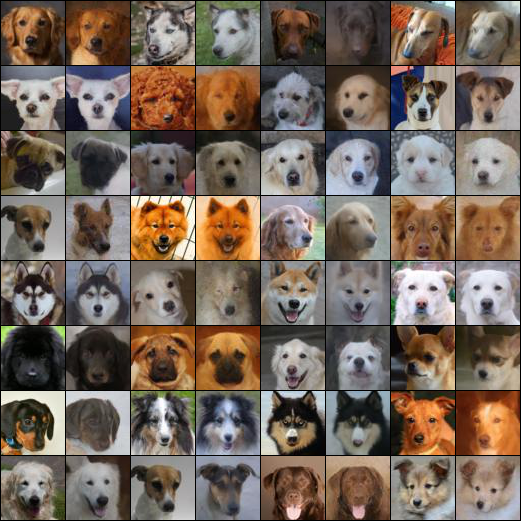}
    \caption{Our Method}
    \label{fig:least}
  \end{subfigure}

  \caption{The similar grid image of the default training and our method. Odd-numbered columns represent images from the training set, while even-numbered columns represent images from the generation set that has the smallest $\ell_{2}$ distance to the corresponding training set image. The images in the grid are arranged in ascending order of $\ell_{2}$ distance, and the selected images from both methods correspond to the same position in terms of $\ell_{2}$ distance.}
  \label{fig:lm}
\end{figure}

\textbf{Visualization.} To provide a more intuitive confirmation of our training method, we visualize the images generated by the model along with their closest counterparts in the training dataset in \cref{fig:lm}.
There is a similarity between the images from the training set and their corresponding images in the generation set, while our method shows lower similarity compared to the default training method.
It is also important to note that our model's FID slightly decreases on all three datasets compared to the default indicating that our method also improves the quality of the generated images.
%This demonstrates a surprising observation that contradicts previous work~\cite{carlini2023extracting} which found that the better diffusion models may be even less private. Therefore, our discovery may significantly aid advancements in future better diffusion models in terms of privacy protection.
% {c|C{0.8cm}C{0.8cm}C{0.8cm}C{0.85cm}|C{0.8cm}C{0.8cm}C{0.8cm}C{0.8cm}|C{0.85cm}C{0.85cm}C{1.0cm}C{0.8cm}}
\begin{table}[tb]
  \centering
  \caption{Comparisons of \crnote{unconditional generation on three datasets} in terms of memorized quantity denoted as MQ. We also report the FID to evaluate the quality of images produced by the model.}% ``-'' represents that the model training has collapsed and there is no result.}
   {\fontsize{5.5}{7.5}\selectfont %
    \begin{tabular}
    {c|ccc|c|ccc|c|ccc|c}
    % {C{1.05cm}|C{0.8cm}C{0.8cm}C{0.95cm}|C{0.55cm}|C{0.9cm}C{0.9cm}C{0.95cm}|C{0.55cm}|C{0.9cm}C{0.9cm}C{0.95cm}|C{0.55cm}}
     \specialrule{\heavyrulewidth}{0pt}{0pt} % 加粗的水平线，位于表格底部
    \hline
    \multirow{2}[4]{*}{Method} & \multicolumn{4}{c|}{CIFAR-10} & \multicolumn{4}{c|}{CIFAR-100} & \multicolumn{4}{c}{AFHQ-DOG} \bigstrut\\
\cline{2-13}          & MQ$_{0.4}$ & MQ$_{0.5}$ & MQ$_{0.6}$$\downarrow$ & FID$\downarrow$   & MQ$_{0.4}$ & MQ$_{0.5}$ & MQ$_{0.6}$$\downarrow$ & FID$\downarrow$   & MQ$_{0.4}$ & MQ$_{0.5}$ & MQ$_{0.6}$$\downarrow$ & FID$\downarrow$ \bigstrut\\

    \hline
    \hline
    Default & 111   & 465   & 2030  & 8.81 & 429   & 1727  & 5620  & 9.29  & 12344 & 19053 & 30795 & 23.59 \bigstrut\\
    
     Adding Noise & 197   & 593   & 2091  & 94.61 & 179   & 1037  & 4383  & 86.18  & 11700 & 19295 & 27224 & 61.18 \bigstrut\\
   
     DP-SGD~\cite{abadi2016deep} & 148   & 728   & 3200  & 12.55 & -   & -  & -  & -  & - & - & - & - \bigstrut\\
     Ambient Diffusion~\cite{daras2024ambient} & 22   & 138   & 851  & 11.7 & -   & -  & -  & -  & - & - & - & - \bigstrut\\
      
      \hline
    IET-AGC & \textbf{14}    & \textbf{117} & \textbf{839}   & \textbf{8.34} & \textbf{144}   & \textbf{760} & \textbf{3274}  & \textbf{8.51} & \textbf{1811}  & \textbf{5435} & \textbf{15237} & \textbf{22.2} \bigstrut\\
    \hline
     \specialrule{\heavyrulewidth}{0pt}{0pt} % 加粗的水平线，位于表格底部
    \end{tabular}%
    }
  \label{tab:attack_result}%
\end{table}%

% \subsection{Fine-tuning Well-trained Diffusion Models}

\label{sec:finetune}
\begin{table}
    \begin{minipage}{0.4\textwidth}
    \setlength{\abovecaptionskip}{-10pt}
    \centering
  \caption{Fine-tuning results of the pre-trained DDPM on CIFAR-10 dataset for 2 epochs.}
  {\fontsize{6.5}{8}\selectfont %
    \begin{tabular}{c|ccc|c}
    \specialrule{\heavyrulewidth}{0pt}{0pt} % 加粗的水平线，位于表格底部
    \hline
    \multirow{2}[4]{*}{Method} & \multicolumn{4}{c}{CIFAR-10} \bigstrut\\
\cline{2-5}          & MQ$_{0.4}$ & MQ$_{0.5}$ & MQ$_{0.6}$$\downarrow$ & FID$\downarrow$ \bigstrut\\
    \hline
    \hline
    Default & 111   & 465   & 2030  & 8.81 \bigstrut\\
    IET  & 51    & 396   & 2367  & 8.33 \bigstrut\\
    AGC & \textbf{44} & \textbf{235} & \textbf{1317} & 11.8 \bigstrut\\
    \hline
    IET-AGC & 83    & 408   & 1796  & \textbf{7.93} \bigstrut\\
    \hline
    \specialrule{\heavyrulewidth}{0pt}{0pt} % 加粗的水平线，位于表格底部
    \end{tabular}%
    }
  \label{tab:finetune_fid}%
    \end{minipage}
    % \hspace{0.05cm}
    \begin{minipage}{0.5\textwidth}
     \centering
  % \vspace{-1.1em}
\caption{Fine-tuning results of Stable Diffusion model on LAION-10k dataset.}
  {\fontsize{6.5}{8}\selectfont %
    \begin{tabular}{p{1.6cm}c|ccc}
     \specialrule{\heavyrulewidth}{0pt}{0pt} % 加粗的水平线，位于表格底部
    \hline
    \multicolumn{2}{c|}{Method} & Sim Score$\downarrow$ & Clip Score$\uparrow$ & FID$\downarrow$ \bigstrut\\
    \hline
    
    \multicolumn{2}{c|}{Default SD} & 0.64  & 30.5  & 18.7 \bigstrut\\
    \hline
    \hline
    \multirow{3}[2]{*}{\makecell{Train Time \\Mitigation~\cite{somepalli2024understanding} }} & MC    & 0.42  & 30.3  & 16.6 \bigstrut[t]\\
          & RC    & 0.57  & 30.6  & \textbf{16.0} \\
          & CWR   & 0.61  & 30.8  & 16.7 \bigstrut[b]\\
    \hline
    \multirow{3}[2]{*}{\makecell{Test Time \\Mitigation~\cite{somepalli2024understanding}}} & RT    & 0.52  & 29.5  & 18.7 \bigstrut[t]\\
          & CWR   & 0.58  & 30.1  & 18.1 \\
          & GNI   & 0.62  & 30.3  & 18.9 \bigstrut[b]\\
    \hline
    \multirow{3}[1]{*}{Our Method} & IET   & 0.41  & \textbf{31.3} & 16.5 \bigstrut[t]\\
          & AGC   & 0.53  & 30.6  & 18.5 \\
          & IET-AGC & \textbf{0.37} & \textbf{31.3} & 16.7 \\
    \hline
    \specialrule{\heavyrulewidth}{0pt}{0pt} % 加粗的水平线，位于表格底部
    \end{tabular}%
    }
  \label{tab:finetnue_SD}%
  % \vspace{-1.4em}
    
    \end{minipage}\hfill
\end{table}

\textbf{Finetuning Unconditional DDPMs.}
% This fine-tuning process only takes two epochs, employing three approaches: IET, AGC, and IET-AGC. 
% The hyperparameters of our three methods remain consistent with those trained from scratch. The results are presented in \cref{tab:finetune_fid}.
\crnote{
% For the CIFAR-10 dataset, we adopt a pre-trained diffusion model and fine-tune it for only two epochs, employing three approaches: IET, AGC, and IET-AGC. The hyperparameters for our three methods remain consistent with those used when training from scratch. 
The results are presented in \cref{tab:finetune_fid}.
}
The IET method shows improvement over the pre-trained in terms of MQ$_{0.4}$ and FID, indicating more efficient data forgetting and higher image quality. The AGC method rapidly causes the model to forget the memorized data. However, it also led to a substantial increase in FID. 
The IET-AGC method not only reduces the FID but also effectively lowers the model's memory rate.

\textbf{Finetuning Text-conditional Stable Diffusion.}
\crnote{
% To further validate our method, we fine-tune the Stable Diffusion model on the LAION-10k dataset using the IET-AGC method for 10 epochs.
% We use the same evaluation metrics as in previous work ~\cite{somepalli2024understanding}: the 95th percentile of similarity scores calculated using object-level SSCD \cite{pizzi2022self} to gauge the degree of memorization by comparing the generation to the original image (Similarity Score), and fidelity to the text condition evaluated with pre-trained CLIP (ViT-B/32) (Clip Score). 
The results are presented in \cref{tab:finetnue_SD}.
Somepalli \etal~\cite{somepalli2024understanding} protects privacy by randomizing conditional information during training and inference, thereby reducing the likelihood of the model replicating specific training data.
Our method IET-AGC achieves the best overall results with a Similarity Score of 0.37, maintaining the highest Clip Score of 31.3 and a competitive FID of 16.7. 
This indicates our approach effectively balances memorization and generative quality, outperforming the default and other mitigation methods.
}

% \begin{figure}[tb]
%   \setlength{\abovecaptionskip}{25pt} % 设置标题上方的间距为 -5pt
%   \setlength{\belowcaptionskip}{-14pt} % 设置标题下方的间距为 -5pt
%   \centering
%   \begin{subfigure}{0.45\linewidth}
%     \includegraphics[width=1.\linewidth]{./imgs/finetune.png}
%     \caption{Without mitigation}
%     \label{fig:Ana_l2_most_least}
%   \end{subfigure}
%   \hspace{0.03\linewidth}
%   \begin{subfigure}{0.45\linewidth}
%     \includegraphics[width=1.\linewidth]{./imgs/finetune_miti.png}
%     \caption{With mitigation using IET-AGC}
%     \label{fig:Ana_spec}
%   \end{subfigure}
%   \caption{ The histograms of top-1 similarity scores between generations and the training data and the top-1 self-similarity scores of training
% data.}
%   \label{fig:Ana_hist}
% \end{figure}

\subsection{Analysis of Skipping}
\label{sec:analysis}
In this section, we conduct comparative experiments on the AFHQ-DOG dataset to delve into which types of images are prone to be skipped, as well as the relationship between memorizable images and those that are skipped.
\begin{figure}[tb]
  \setlength{\abovecaptionskip}{25pt} % 设置标题上方的间距为 -5pt
  \setlength{\belowcaptionskip}{-14pt} % 设置标题下方的间距为 -5pt
  \centering
  \begin{subfigure}{0.45\linewidth}
    \includegraphics[width=1.\linewidth]{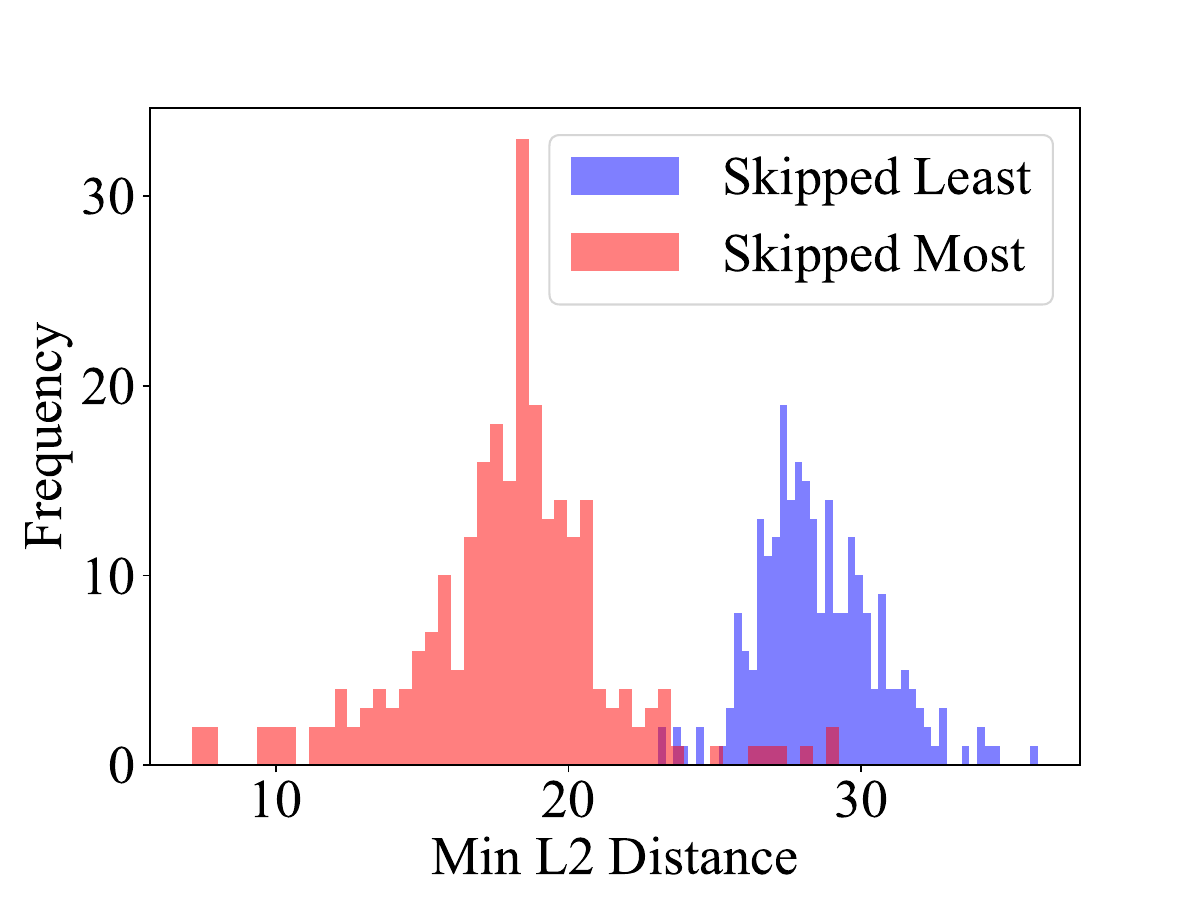}
    \caption{Distribution of distances to the most similar images in the dataset.}
    \label{fig:Ana_l2_most_least}
  \end{subfigure}
  % \hspace{0.03\linewidth}
  \begin{subfigure}{0.45\linewidth}
    \includegraphics[width=1.\linewidth]{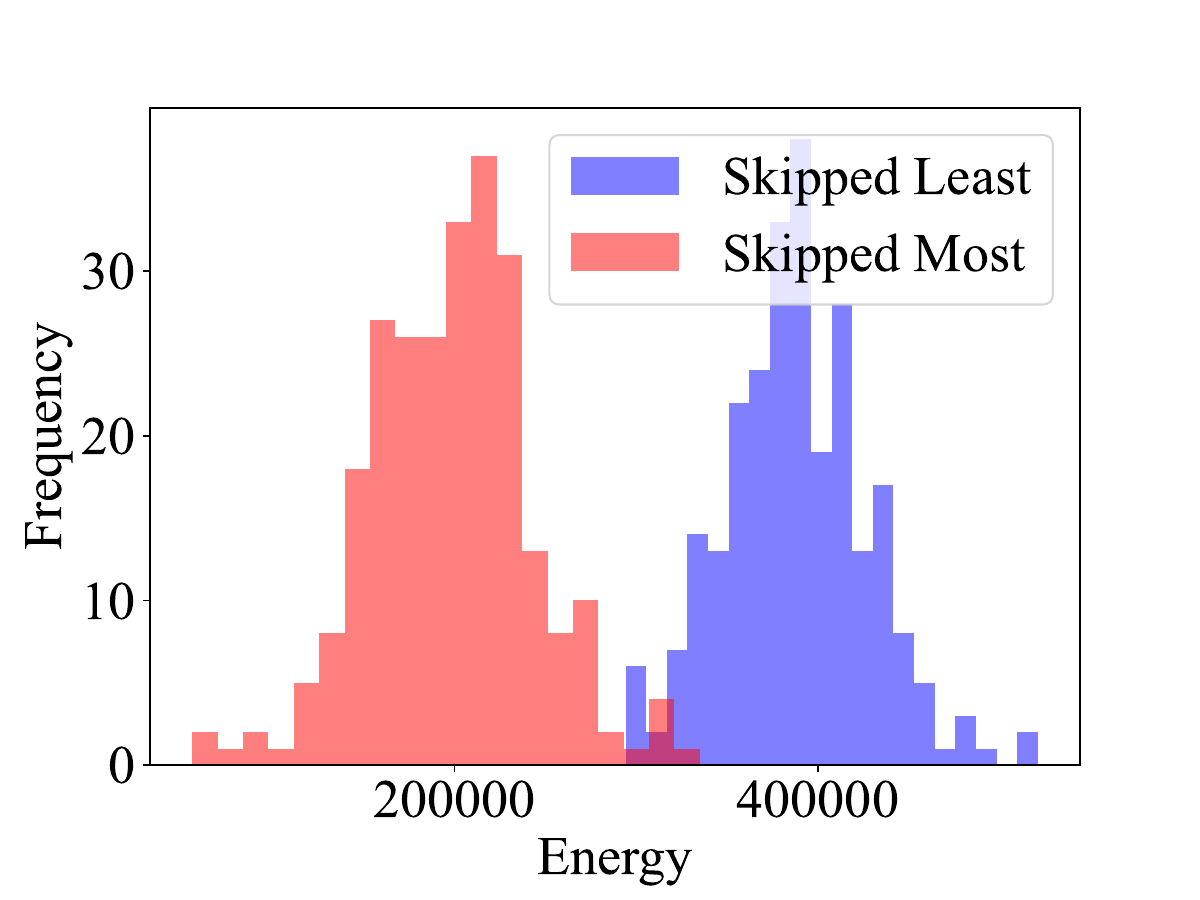}
    \caption{Energy distribution. The greater the energy, the more complex the image.}
    \label{fig:Ana_spec}
  \end{subfigure}
  \caption{ Most skipped images vs least skipped images.}
  \label{fig:Ana_hist}
\end{figure}
\subsubsection{Frequency of Skipped Images.}
\begin{figure}[tb]
  \setlength{\abovecaptionskip}{1pt} % 设置标题上方的间距为 -5pt
  \setlength{\belowcaptionskip}{-13pt} % 设置标题下方的间距为 -5pt
  \centering
  \includegraphics[height=6.5cm]{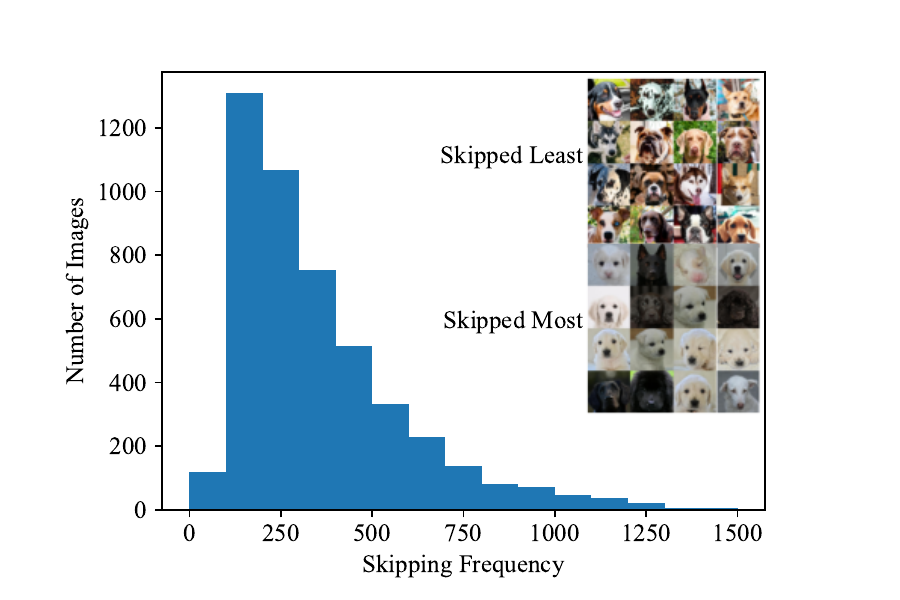}
  \caption{
    Distribution of skipped image counts. 
  }
  \label{fig:value_counts}
\end{figure}

Throughout the training process, we record the identifiers of skipped images. 
As shown in \cref{fig:value_counts}, our method does not entail skipping all images. 
In our approach, 90\% of the images are skipped fewer than 648 times (across a total of 2,278 training epochs), indicating that our method can effectively differentiate between different images. 
This suggests that we are not simply reducing memorization by constraining the model's learning. 
On the other hand, while our method requires skipping images with exceptionally low loss values, all images still contribute to the model's training.
% We also display the images most frequently skipped and those least likely to be skipped.

\subsubsection{Images Most Easily Skipped.}
We believe these images are more easily skipped for two main reasons.
Firstly, data aggregation. 
We computed the $\ell_{2}$ distance between these easily skipped images and all other images in the dataset, as well as between those not easily skipped images and all other images in the dataset.
\cref{fig:Ana_l2_most_least} indicates that the distribution of the skipped images is more clustered.
Consistent with the findings of Carlini \etal~\cite{carlini2023extracting}, which suggested that removing duplicate training images effectively reduces memorization capacity, skipping these clustered images can also reduce memorization capacity.
Secondly, data simplicity. 
We performed Fourier transforms on these easily skipped and not easily skipped images to obtain their energy distributions, as shown in \cref{fig:Ana_spec}. 
The easily skipped images have less energy, indicating that they lack finer details.
We believe both factors contribute to making these images easier for the model to memorize, thus skipping them has a positive impact on reducing the model's memorization capacity.

\subsection{Ablation Study}
\label{sec:ablation}
\subsubsection{Performance Comparisons of Each Component.}
To further understand the effectiveness of our approach, we conduct ablation experiments to investigate the individual impacts of different components on CIFAR-10. In addition, in our approach, the dataset is evenly distributed, meaning each data shard is set in an IID (independently and identically distributed) manner. To validate the reasonableness of this dataset configuration, we also establish an experiment where each data shard is set up in a non-independent and identically distributed (non-IID) manner. Similar to~\cite{hsu2019measuring}, we employ the Dirichlet distribution to generate data, thereby establishing such a setting.

\begin{wraptable}{r}{0.4\textwidth}
    \setlength{\abovecaptionskip}{-8pt}
  \caption{Performance comparisons of each component.}
  {\fontsize{6.5}{8}\selectfont %
    \begin{tabular}{c|ccc|c}
     \specialrule{\heavyrulewidth}{0pt}{0pt} % 加粗的水平线，位于表格底部
    \hline
    \multirow{2}[4]{*}{Method} & \multicolumn{4}{c}{CIFAR-10} \bigstrut\\
\cline{2-5}          & MQ$_{0.4}$ & MQ$_{0.5}$ & MQ$_{0.6}$$\downarrow$ & FID$\downarrow$ \bigstrut\\
    \hline
    \hline
    Default & 111   & 465   & 2030  & 8.81 \bigstrut\\
   
    IET$_\mathrm{IID}$  & 89    & 382   & 1783  & \textbf{7.61} \bigstrut\\
    IET$_\mathrm{non-IID}$ & 69    & 315   & 1439  & 12.5 \bigstrut\\
    
    AGC & 26    & 154   & 976   & 11.36 \bigstrut\\
    \hline
    IET-AGC & \textbf{14} & \textbf{117} & \textbf{839} & 8.34 \bigstrut\\
    \hline
    \specialrule{\heavyrulewidth}{0pt}{0pt} % 加粗的水平线，位于表格底部
    \end{tabular}%
    }
  \label{tab:ablation_cifar10}%
  % \vspace{-1cm}
\end{wraptable}

Results in ~\cref{tab:ablation_cifar10} show that when the data shards are set in non-IID, Iterative Ensemble Training (IET) successfully reduces the memorization (IET$_\mathrm{non-IID}$).
However, the non-IID splitting reduces the quality of performance since the divergent optimization of different models and aggregating these models affects the performance. 
Differently, when the data shards are set in IID,  IET$_\mathrm{IID}$ can reduce both the memorization and FID score. 
Thus in this paper, we choose to use the IID splitting strategy. 
Besides, ~\cref{tab:ablation_cifar10} shows that both Anti-Gradient Control (AGC) and Iterative Ensemble Training (IET) reduce memorization effectively and AGC is more effective than the IET on the memorization reduction.
Compared to the conventional method, AGC reduces MQ$_{0.5}$ by approximately 66\%, while IET$_\mathrm{IID}$ reduces it by around 18\%. 
However, using AGC only shows a slight growth in FID, which affects the quality of the generated images.
Interestingly, combining both IET and AGC leads to better FID and memory reduction effects.
% When data shards are set in non-IID, Each component has the ability to reduce memory. 
% AGC is more effective than the IET. 
% Compared to the conventional method, AGC reduces MQ$_{0.5}$ by approximately 66\%, while IET reduces it by around 18\%. 
% However, IET shows a decrease in FID, even lower than our method, while AGC shows a slight improvement in FID. 
% Interestingly, combining both components leads to better FID and memory reduction effects.

\subsubsection{Exploring Parameters Impact on Experimental Results.}
\label{sec:impact}
In this study, we examine how various parameters affect our experimental outcomes.
By systematically varying these parameters, we aim to understand how they influence our results and to identify the optimal settings for our experiments. 
Specifically, we conduct a series of experiments where we change the number of shards, training epochs per aggregation period, skipping threshold $\lambda$, and smoothing factor $\gamma$ of the memory bank. For each variation, we measure the impact on MQ and FID. The default parameters are set with the number of shards $K$ as 10, epochs per shard $E$ as 100, skipping threshold $\lambda$ as 0.5, and smoothing factor $\gamma$ as 0.8. 
Results are reported in \cref{tab:parameters_effect}.

\textbf{Number of Shards $K$.} We investigate the impact of the number of data shards on model performance by setting it to 1, 5, 10, and 15. $K=1$ means the default training strategy of diffusion models. Results show that the MQ scores of $K=5,10,15$ are all lower than $K=1$, indicating that using our IET method can effectively reduce memorization. When $K=10$, the MQ score achieves the best performance. However, with the growth of $K$, the FID score increases. Thus we need to balance the FID and MQ to choose the proper number of shards.
%The experiment reveals a general trend: as the value of this parameter escalates, FID and MQ demonstrate a significant increase. When the shard is set as 15, 

\textbf{Training Epochs per Aggregation Period $E$.} 
We set the total number of training epochs unchanged and conduct experiments by varying the number of epochs per model aggregation period, \ie, the aggregating frequency of IET. Results show that both high and low frequencies of aggregation will reduce the performance of the memorization and image quality. Thus, we choose $E=100$ to achieve the best performance of MQ.

\begin{wraptable}{r}{0.6\textwidth}
  \setlength{\abovecaptionskip}{-13pt} 
  \caption{Parameters impact on experimental results.}
  {
  \fontsize{6.5}{8}\selectfont %
    \begin{tabular}{cc|ccc|c}
    \specialrule{\heavyrulewidth}{0pt}{0pt} % 加粗的水平线，位于表格底部
    \hline
    \multicolumn{2}{c|}{\multirow{2}[3]{*}{Parameters}} & \multicolumn{4}{c}{CIFAR-10} \bigstrut[b]\\
\cline{3-6}    \multicolumn{2}{c|}{} & MQ$_{0.4}$ & MQ$_{0.5}$ & MQ$_{0.6}$$\downarrow$ & FID$\downarrow$ \bigstrut\\
    \hline
    \hline
    \multirow{3}[2]{*}{Number of Shards $K$} 
    
    & 1    & 111 & 465 & 2030   & 8.81 \bigstrut[t]\\
     & 5     & 20    & 129   & \textbf{798} & \textbf{6.37}\\
     & 10    & \textbf{14} & \textbf{117} & 839   & 8.34 \\
        & 15    & 21    & 169   & 1143  & 9.1 \bigstrut[b]\\

    \hline
    \multirow{5}[0]{*}{Epoches per Aggregation $E$} & 10    & 26    & 167   & 1018  & 9.31 \\
          & 50    & 16    & 194   & 1120  & \textbf{7.76} \\
    & 100   & \textbf{14} & \textbf{117} & \textbf{839} & 8.34 \\
    & 200    & 17    & 193   & 1307  & 9.01 \\
    & 250    & 24    & 181   & 1114  & 10.36 \\
    \hline
    \multirow{4}[1]{*}{Skipping Threshold ${\lambda}$}  & 0.4   & 47    & 310   & 1573  & \textbf{7.72} \\
     & 0.5     & 14    & 117   & 839   & 8.34 \\
     & 0.66   & \textbf{3} & 69 & 608 & 8.25 \\
     & 0.8   & \textbf{3}    & \textbf{45}   & \textbf{503}  & 10.43 \\
    \hline
    \multirow{3}[2]{*}{Smoothing Factor $\gamma$} & 0.5   & 27    & 192   & 1193  & 8.99 \bigstrut[t]\\
     & 0.8   & \textbf{14} & \textbf{117} & \textbf{839} & \textbf{8.34} \\
          & 0.9   & 20    & 132   & 886   & 8.94 \bigstrut[b]\\
    \hline
    \specialrule{\heavyrulewidth}{0pt}{0pt} % 加粗的水平线，位于表格底部
    
    \end{tabular}%
    }
  \label{tab:parameters_effect}%
  % \vspace{-2cm}
\end{wraptable}%

% We conduct experiments by varying the number of epochs per data shard, setting E to 10, 50, 100,200, and 250. As we can see, when the epochs per data shard are set as 100 or 50, the model has the weakest memory capacity or the lowest FID. As the epoch gradually increases beyond 100 or decreases below 100, the model's memory capacity also progressively strengthens.

\textbf{Skipping Threshold $\lambda$.} We evaluate the importance of $\lambda$ in mitigating the memorization effect by setting the values of $\lambda$ to $0.4$, $0.5$, $0.66$ and $0.8$. A large threshold means skipping more training samples that are easily memorized. Results in ~\cref{tab:parameters_effect} show as $\lambda$ grows, more memorable training samples are skipped and the memorization phenomenon is further reduced. However, skipping more samples will reduce the model performance, \ie, the generation quality. 

% When $\lambda$ is set to $0.4$, the FID increases only by $1.44\%$ compared to iterative ensemble training. However, there is a decrease of $47.1\%$ in the MQ$_{0.4}$ metric, which reflects the effectiveness of our method. When we set the value of $\lambda$ to $0.66$, all three metrics, MQ$_{0.4}$ MQ$_{0.5}$ and MQ$_{0.6}$, show further decreases.

\textbf{Smoothing Factor $\gamma$.} The selection of the smoothing factor $\gamma$ is also crucial. When we set $\gamma$ to 0.9, the update of the memory bank becomes overly sluggish, failing to faithfully reflect the current model's loss across different time steps. Conversely, if we set $\gamma$ to 0.5, the memory bank becomes excessively sensitive, which can lead to instability and fragility during the training process.

% {c|C{1.0cm}C{1.0cm}C{1.0cm}C{1.0cm}}   ablation_cifar10


\section{Limitations and Future Work}
Although our strategy has proven effective in reducing the model's memorization capacity, the selection of the threshold $\lambda$, which balances between the model's memorization capacity and the quality of the generated images, can only rely on experiments. 
% This means that the optimal value of $\lambda$ must be determined empirically through extensive testing and evaluation on different datasets and scenarios. 
% The process can be time-consuming and resource-intensive, as it involves iterating through various values and assessing their impact on both memorization and image quality.
In future research, we plan to design a quick analysis method for datasets to help determine the appropriate choice of threshold.

\section{Conclusion}
This paper innovatively proposes a training strategy for diffusion models, which train models on multiple data shards and ignore data with abnormal loss values. 
This strategy effectively reduces the memory capacity of the model, further strengthening data privacy protection without compromising the quality of the generated images. 
We firmly believe that this training strategy has a broad application prospect and great development potential in the field of data privacy protection.

% \clearpage
\crnote{
\section*{Acknowledgments}
This work was supported by National Natural Science Foundation of China (Grant No. 62306273), National Natural Science Foundation of China (Grant No. 62372341), and the Fundamental Research Funds for the Central Universities (Grant No. 2042024kf0040).
The numerical calculations in this paper have been done on the supercomputing system in the Supercomputing Center of Wuhan University.
}

\bibliographystyle{splncs04}
\bibliography{main}
\clearpage
\input{appendix.tex}

\end{document}

%% file: introduction.tex
\jxnote{Recent advancements in diffusion models have significantly transformed the landscape of image generation. Modern diffusion models, such as Stable Diffusion~\cite{rombach2022high}, Midjourney~\cite{midjourney2022}, and SORA~\cite{sora2024}, have the ability to generate realistic images that are hard for humans to distinguish, demonstrating the unparalleled capabilities in producing diverse images. However, recent works~\cite{carlini2023extracting,somepalli2024understanding,wen2023detecting} suggested that diffusion models are capable of memorizing images from the training set and reproducing them, which has the potential risk of privacy leakage. To address the severe problem, some works~\cite{zhang2023forget,ni2023degeneration,gandikota2024unified,kumari2023ablating} proposed to make diffusion models ``forget'' specific concepts such as a portrait of a certain celebrity, or the style of a particular artist. However, these works can only blacklist specific content that users want to conceal, but cannot completely cover the privacy-sensitive information that the model might remember, which still has a risk of privacy leakage.}

\jxnote{Recently, some works~\cite{somepalli2023diffusion,daras2024ambient,somepalli2024understanding,wen2023detecting} have proposed to mitigate diffusion memorization without specific content limitations, thus reducing the risk of diffusion models leaking privacy-sensitive training data. Most of them focused on tackling the training data memorization in text-to-image diffusion models, and proposed data augmentation for captions to reduce model memorization since the insufficient diversity in captions easily leads to training data generation. For instance, Somepalli \etal~\cite{somepalli2024understanding} utilized random caption replacement, random token replacement, caption word repetition, \etc, to reduce memorization. Another previous approach Daras \etal~\cite{daras2024ambient} proposed to use corrupted images for diffusion model training for memorization reduction.}  \crnote{Wen \etal~\cite{wen2023detecting} introduced methods for detecting memorized prompts through text-conditional predictions and proposed two strategies to mitigate memorization: minimizing during inference or filtering during training.}

\crnote{Although these works represented an important step forward in understanding the memorization issue in diffusion models, they either utilized simple data augmentation strategies or only focused on the easily memorable images that are related to specific captions in cross-modal generation tasks. However, diffusion models have been proven capable of generating images from memory without the need for text guidance~\cite{carlini2021extracting}, and previous methods focused on a limited scope of the memorization issue in diffusion models.
%These approaches were not necessarily capable of addressing the model's core ability to mitigate memory and 
%could not handle generation tasks without the text modality. 
Differently, in this paper, we propose a novel training framework for diffusion models from the perspective of visual modality, i.e., IET-AGC (Iterative Ensemble Training with Anti-Gradient Control), which not only reduces memorization fundamentally but also provides a more generic approach for both unconditional and text-conditional diffusion models.}

\jxnote{First, we propose an iterative ensemble training (IET) framework to mitigate memorization by parameter aggregation. Training data are stored in parameters of diffusion models due to over-optimization, and the model ensemble strategy aggregates parameters to re-organize model knowledge, which contributes to alleviating the memorization of training data. Thus, we separate the training data into several groups, train diffusion models individually, and ensemble them for memorization reduction. However, aggregating models trained on subsets of data may potentially decrease the performance due to the divergent optimizations on different subsets of these models. Motivated by Federated Learning techniques~\cite{mcmahan2017communication}, we iteratively ensemble the models during training, which reduces memorization by multiple aggregation and maintains generation performance.}

\jxnote{Based on our IET training framework, we further introduce an Anti-Gradient Control module to reduce memorization of training data ulteriorly. We conduct practical analysis on losses during training and find that the training loss for easily memorable images tends to be obviously lower than that for less memorable images. Thus, we propose to exclude the sample with a relatively small loss value from the current mini-batch to avoid memorizing these samples. During the training process, as the diffusion model exhibits varying average loss values across different time steps, we maintain a memory bank to store the average loss of each time step. Subsequently, we discard samples with losses below a certain proportion of the average loss. Since the model has encountered the sample during training, excluding it is unlikely to have a significant impact on the model's performance.}

\crnote{Extensive experiments on four datasets show the significance of our IET-AGC framework. 
Our method significantly reduces the memorized quantity by 87.3\%, 66.4\%, and 85.3\% compared with the default training (DDPM~\cite{ho2020denoising}) on CIFAR-10 and CIFAR-100 and AFHQ-DOG, respectively.
In addition, considering the high cost of re-training existing well-trained diffusion models, we also propose an efficient way to address the memorization issue of existing pre-trained models by simply fine-tuning using our method for several epochs.
% Experiments show that continuing fine-tuning with our method for only two epochs can significantly reduce the model's memorization of training data by 25.22\% on CIFAR-10 and , illustrating the effectiveness and applicability.
Experiments show that using our method to fine-tune for just two epochs can significantly reduce the memorization by 25.22\% in unconditional diffusion models on CIFAR-10.
Furthermore, when fine-tuning the text-conditional diffusion model, Stable Diffusion, our approach decreases the memorization score by 42.18\% compared to conventional fine-tuning methods.
These results demonstrate that our method performs excellently on both unconditional and text-conditional diffusion models.
}

% \jxnote{Our contributions can be summarized as follows:}

% \jxnote{}

%% file: appendix.tex
% \documentclass[runningheads]{llncs}

% ---------------------------------------------------------------
% Include basic ECCV package
 
% TODO REVIEW: Insert your submission number below by replacing '*****'
% TODO FINAL: Comment out the following line for the camera-ready version
% \usepackage[review,year=2024,ID=11800]{eccv}
% TODO FINAL: Un-comment the following line for the camera-ready version
% \usepackage{eccv}

% OPTIONAL: Un-comment the following line for a version which is easier to read
% on small portrait-orientation screens (e.g., mobile phones, or beside other windows)
%\usepackage[mobile]{eccv}

% ---------------------------------------------------------------
% Other packages

% Commonly used abbreviations (\eg, \ie, \etc, \cf, \etal, etc.)
% \usepackage{eccvabbrv}

% Include other packages here, before hyperref.

\appendix

\section{Algorithm}
\begin{algorithm}[!h]
\label{algo}
\caption{The IET-AGC Framework}
\SetAlgoLined
\LinesNotNumbered
\KwIn{Dataset $D$, training rounds $M$, training epochs per aggregation period $E$, number of shards $K$, initial model $\theta^0$, skipping threshold $\lambda$, smoothing factor $\gamma$, memory bank $l$, learning rate $\eta$ }
\KwOut{Diffusion model $\theta^M$}

Divide dataset $D$ into an equal number of shards $D_i$, $i\in [1,...K]$ \\
Initialize memory bank $l$ with all elements as $0$ \\

\For{$m=1$ \KwTo $M$}{
    \For{$i=1$ \KwTo $K$}{
        Initialize model $\theta_{i}^m \leftarrow \theta^{m-1}$ \\
        \For{$e=1$ \KwTo $E$}{            
            $x \sim D_i, \epsilon  \sim \mathit{N} (0, I) $ \\
            $t$ $\sim$ Uniform $({1, ..., T})$\\        
            $Loss = \mathcal{L} (x, t, \epsilon; \theta_{i}^m)$ \\
            \If{$\frac{Loss}{l_t} < \lambda$}
            {
                % \vspace{0.05cm}
                $Loss \leftarrow 0$
            }
            $l_{t} \leftarrow \gamma \cdot l_t + (1 - \gamma) \cdot \mathcal{L} (x, t, \epsilon; \theta_{i}^m)$ \\
            $\theta_{i}^m \leftarrow \theta_{i}^m - \eta \nabla Loss$
        }
    }
    $\theta^m\leftarrow \frac{1}{K}\sum_{i=1}^{K}\theta_{i}^m$
}
\end{algorithm}

\section{Experiments of DP-SGD with varying noise multipliers}
We conduct a series of experiments with DP-SGD (Differentially-Private Stochastic Gradient Descent)~\cite{abadi2016deep} changing the noise multipliers to 0.0002, 0.0005, and 0.0008 to compare our method with different noise levels in DP. Results are shown in ~\cref{tab:dp_parameters}.  When the noise multiplier is set to 0.0005, DP-SGD achieves the best scores in terms of MQ and FID. However, all DP-SGD results improve the FID score compared with the baseline model. When $\tau$=0.0005, DP-SGD slightly reduces the memorization (101 v.s. 111), but it is still far from comparable to the memorization reduction capability of our method.
%our method still surpasses it, demonstrating lower memory usage and superior image quality.
% \vspace{-0.5cm}
\begin{table}[h]
  \centering
  \caption{Performance of DP-SGD across multiple experiments with varying noise multiplier $\tau$.}
  {\fontsize{6.5}{8}\selectfont %
    \begin{tabular}{c|ccc|c}
    \specialrule{\heavyrulewidth}{0pt}{0pt} % 加粗的水平线，位于表格底部
    \hline
    \multirow{2}[4]{*}{Method} & \multicolumn{4}{c}{CIFAR-10} \bigstrut\\
\cline{2-5}         & MQ$_{0.4}$ & MQ$_{0.5}$ & MQ$_{0.6}$$\downarrow$ & FID$\downarrow$  \bigstrut\\
    \hline
    \hline
    Default & 111   & 465   & 2030  & 8.81 \bigstrut[t]\\
    DP-SGD~\cite{abadi2016deep} $\tau$=0.0002 & 148   & 728   & 3200  & 12.55 \\
    DP-SGD~\cite{abadi2016deep} $\tau$=0.0005 & 101   & 380   & 1716  & 10.02 \\
    DP-SGD~\cite{abadi2016deep} $\tau$=0.0008 & 124   & 549   & 2498  & 13.82 \bigstrut[b]\\
    \hline
    IET-AGC & \textbf{14} & \textbf{117} & \textbf{839} & \textbf{8.34} \bigstrut\\
    \hline
     \specialrule{\heavyrulewidth}{0pt}{0pt} % 加粗的水平线，位于表格底部
    \end{tabular}%
    }
  \label{tab:dp_parameters}%
\end{table}%
\section{More Ablation Study}
\crnote{We randomly split the data evenly on AFHQ-DOG and LAION-10k, where no class label is available on these datasets.
Experiments demonstrate the effectiveness of our method in this setting.
% Specifically, on AFHQ, our method resulted in a 71\% decrease in MQ0.5 (5435 vs 19053) and a 5.8\% decrease in FID (22.2 vs 23.6).
Additionally, we conduct ablation experiments on CIFAR-10 by randomly splitting the data evenly (without using class labels). Results are shown in ~\cref{tab:nonclass}.
% We believed that a uniformly distributed class was better for the model’s performance.
We find randomly splitting (without class labels) slightly affects the generation quality. }
% \begin{table}[htpb]
%   \centering
%   \caption{Ablation experiments of randomly splitting the data evenly.}
%   \vspace{-1.1em}
%   \setlength{\abovecaptionskip}{-40pt}
%     {\fontsize{6.5}{8}\selectfont %
%     \begin{tabular}{P{4.0cm}|P{1.0cm}P{0.8cm}}
%      \specialrule{\heavyrulewidth}{0pt}{0pt} % 加粗的水平线，位于表格底部
%     \hline
%     \multirow{2}[4]{*}{Method} & \multicolumn{2}{c}{CIFAR-10}  \bigstrut\\
% \cline{2-3}          & MQ$_{0.5}$$\downarrow$ & FID$\downarrow$   \bigstrut\\
%     \hline
%     \hline
%     Default  & 465   & 8.81 \bigstrut[t]\\
     
%     IET-AGC (w/o class label) & \textbf{91} & 9.12    \\
%     IET-AGC (w/ class label) & 117   & \textbf{8.34}  \bigstrut[b]\\
%     \hline
%     \specialrule{\heavyrulewidth}{0pt}{0pt} % 加粗的水平线，位于表格底部
%     \end{tabular}%
    
%     }
%     \label{tab:nonclass}%
%     \vspace{-1.6em}
% \end{table}%
\begin{table}[t]
  \centering
  \caption{Ablation experiments of randomly splitting the data evenly.}
  % \vspace{-1.1em}
  % \setlength{\abovecaptionskip}{-40pt}
    {\fontsize{6.5}{8}\selectfont %
    \begin{tabular}{C{4.0cm}|C{1.0cm}C{1.0cm}C{1.0cm}|C{0.8cm}}
     \specialrule{\heavyrulewidth}{0pt}{0pt} % 加粗的水平线，位于表格底部
    \hline
    \multirow{2}[4]{*}{Method} & \multicolumn{4}{c}{CIFAR-10}  \bigstrut\\
\cline{2-5}        & MQ$_{0.4}$  & MQ$_{0.5}$ & MQ$_{0.6}$$\downarrow$ & FID$\downarrow$   \bigstrut\\
    \hline
    \hline
    Default & 111 & 465   & 2030  & 8.81 \bigstrut\\
     
    IET-AGC (w/o class label) & \textbf{10} & \textbf{91} & \textbf{769} & 9.12  \bigstrut\\
    IET-AGC (w/ class label)& 14   & 117  & 839 & \textbf{8.34}  \bigstrut\\
    \hline
    \specialrule{\heavyrulewidth}{0pt}{0pt} % 加粗的水平线，位于表格底部
    \end{tabular}%
    
    }
    \label{tab:nonclass}%
    % \vspace{-1.6em}
\end{table}%
% We show the pseudo-code for our proposed IET-AGC approach in Algo.~\cref{algo}
% \vspace{-1.0cm}

\section{Implementational Details}
\crnote{
% \jxnote{We employ the model architecture of DDPM~\cite{ho2020denoising} which constructs using the U-Net architecture~\cite{ronneberger2015u}. 
% We keep the conventional training method of DDPM~\cite{ho2020denoising} and IET-AGC in the same settings to verify our method. 
%According to Gu \etal~\cite{gu2023memorization}, many factors, to varying degrees, influence the memory capacity of the Diffusion model, especially data dimension, model size, time embedding, and class conditions. Therefore, in both methods, we set parameters consistently, including model configuration, batch size, training iterations, and so on.
When conducting experiments on training Diffusion models from scratch using CIFAR-10 and CIFAR-100, we set the batch size to 128 and train for 400k and 580k iterations, respectively. 
On AFHQ-DOG, the batch size is set to 60, and we train for 180k iterations. 
In the IET framework, CIFAR-10 and CIFAR-100 are divided into ten shards, each containing the same number of classes and instances.
AFHQ-DOG is divided into five shards based on the number of instances, as it lacks class information. 
On the CIFAR-10 and CIFAR-100 datasets, we set the threshold $\lambda$ to $0.5$, indicating that data with loss less than half of the average loss is skipped. 
For the AFHQ-DOG dataset, due to its smaller size and pronounced memory phenomena, we adjust the threshold $\lambda$ to $0.714$.
We also perform experiments on fine-tuning pre-trained Diffusion models using the CIFAR-10 dataset, maintaining the same hyperparameters as those used for training from scratch, but only for 2 epochs. 
To demonstrate the effectiveness of our method in text-conditioned Diffusion models, we fine-tune Stable Diffusion on LAION-10k. 
The IET framework divides the LAION-10k dataset into 8 shards, with the threshold $\lambda$ set to 0.8. 
We set the batch size to 8 and fine-tune Stable Diffusion for 200k iterations.
On all datasets, the smoothing factor $\gamma$ for the memory bank is set to 0.8. 
}

\section{Loss Analysis on CIFAR-100 and AFHQ-DOG}
\begin{figure}
  \centering
  \setlength{\abovecaptionskip}{-7pt} % 设置标题上方的间距为 -5pt
  \setlength{\belowcaptionskip}{-48pt} % 设置标题下方的间距为 -5pt
  \includegraphics[height=6.5cm]{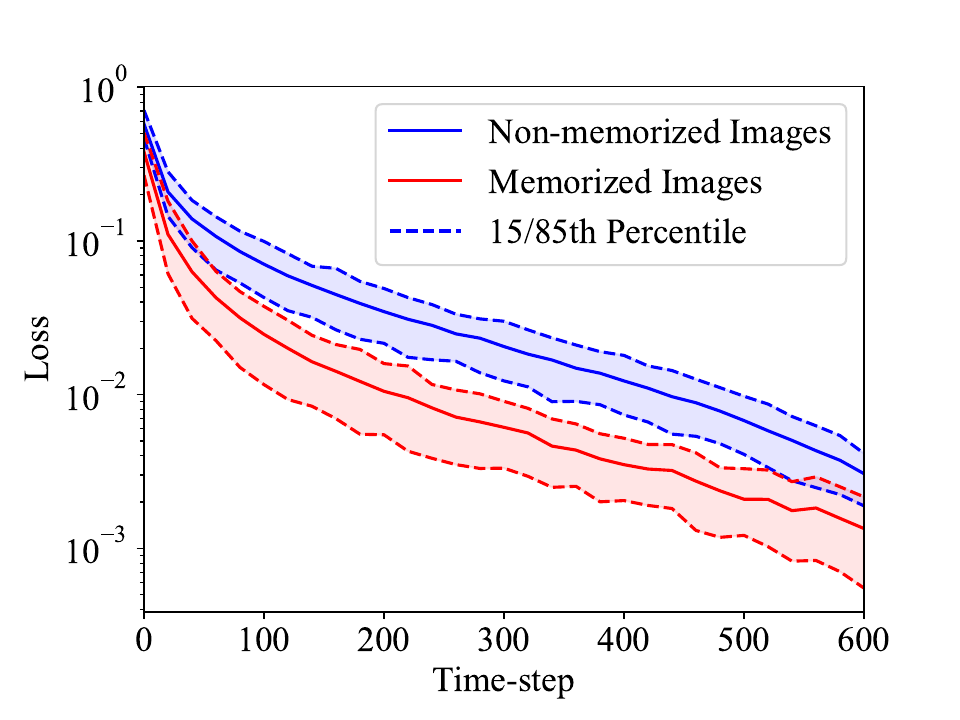}
  \caption{
  Comparison of the losses between memorized and non-memorized images on CIFAR-100. The solid line represents the averaged losses of memorized and non-memorized images, while the two dashed lines represent the losses of the 15th and 85th percentile data respectively.
  }
  \label{fig:cifar100_loss}
\end{figure}

\begin{figure}[h]
  \centering
  \setlength{\abovecaptionskip}{-8pt} % 设置标题上方的间距为 -5pt
  \setlength{\belowcaptionskip}{-30pt} % 设置标题下方的间距为 -5pt
  \includegraphics[height=6.5cm]{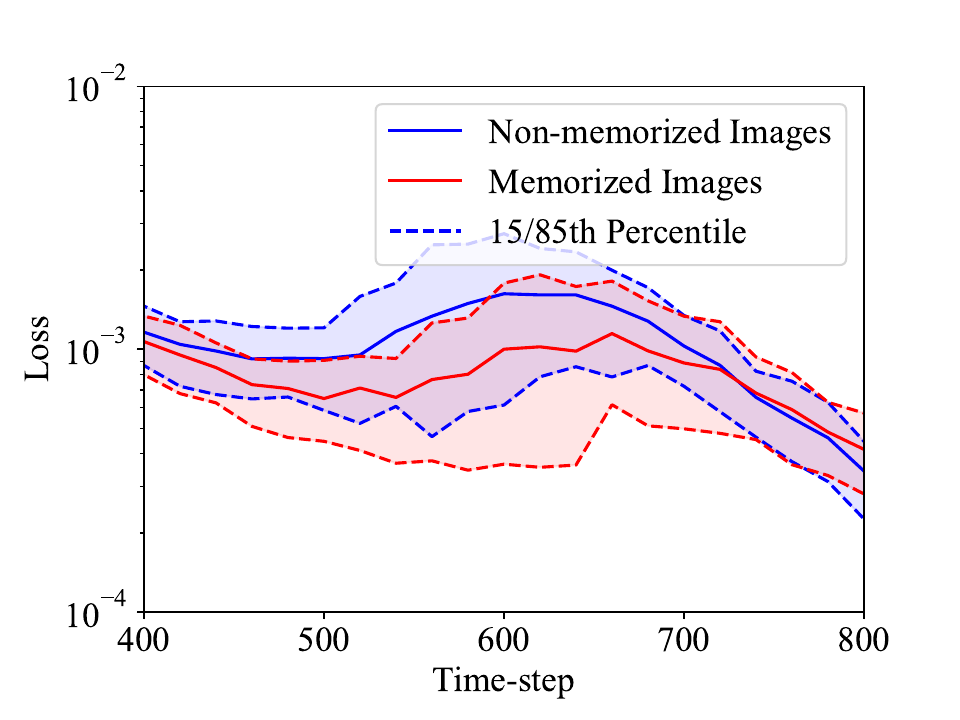}
  \caption{
  Comparison of the losses between memorized and non-memorized images on AFHQ-DOG. The solid line represents the averaged losses of memorized and non-memorized images, while the two dashed lines represent the losses of the 15th and 85th percentile data respectively.
  }
  \label{fig:afhq_loss}
\end{figure}

We present the results of loss analysis for CIFAR-100 and AFHQ-DOG in \cref{fig:cifar100_loss} and \cref{fig:afhq_loss}. 
The results obtained on CIFAR-100 show similarities to those on CIFAR-10. 
However, on AFHQ-DOG, due to its fewer images, the model exhibits a memorization phenomenon across the entire dataset, resulting in less noticeable differences.
% \vspace{-0.7cm}
% \clearpage
% \vspace*{0pt}
% \vspace{-1.0cm}

% \vspace{-1.0cm}

% \begin{algorithm}[!h]
% \SetAlgoLined
% \LinesNotNumbered
% \caption{The IET-AGC Framework}

% \KwIn{Dataset $D$, training rounds $M$, training epochs per aggregation Period $E$, Initial model $\theta^0$, skipping threshold $\lambda$, smoothing factor $\gamma$, time-step $t\in [0,T]$, memory bank at time-step $t$ $l_{t}$, gaussian noise vector $\epsilon$}
% \KwOut{$\theta^M$}
% divide the dataset $D$ into equal shards according to \cref{eq:divided_datasets}\\
% \For{$m$ = 1 to $M$}{
%     \For{each shard $i$}{
%         \State Initialize model $\theta_{i,0}^m = \theta^{m-1}$\\
%         \For{$e$ = 1 to $E$}{
            
%             % \State Train single Diffusion model $\theta_{i,e}^m$ $\leftarrow$ 
            
%             Compute loss $\mathcal{L} (x, t, \epsilon; \theta_{i,e}^m) \leftarrow CE(\epsilon,\epsilon_{\theta_{i,e}^m})$ in ~\cref{eq:loss} \\
%              Get $mask \leftarrow Compare(\mathcal{L} (x, t, \epsilon; \theta_{i,e}^m), l_{t},\lambda)$ in ~\cref{eq:ratio} and in ~\cref{eq:update_loss} \\
%             Update loss $\mathcal{L} (x, t, \epsilon; \theta_{i,e}^m) \leftarrow \mathcal{L} (x, t, \epsilon; \theta_{i,e}^m) \odot mask$ in ~\cref{eq:update_loss} \\
%              Update $l_{t} \leftarrow EMA(\gamma,l_{t},\mathcal{L} (x, t, \epsilon; \theta_{i,e}^m))$ in ~\cref{eq:ema} \\
%             Get model $\theta_{i,e}^m$ according to ~\cref{eq:local_update}
%         }
%         \State Set $\theta_i^m = \theta_{i,E}^m$
%     }
%     \Statex Get merged model $\theta^{m}$ according to \cref{eq:average}
% }
% % \Return{$\theta^M$}

% \end{algorithm}